\documentclass{article}

\usepackage{arxiv}
\usepackage[utf8]{inputenc} % allow utf-8 input
\usepackage[T1]{fontenc}    % use 8-bit T1 fonts
\usepackage{hyperref}       % hyperlinks
\usepackage{url}            % simple URL typesetting
\usepackage{booktabs}       % professional-quality tables
\usepackage{amsfonts}       % blackboard math symbols
\usepackage{nicefrac}       % compact symbols for 1/2, etc.
\usepackage{microtype}      % microtypography
\usepackage{lipsum}
\usepackage{multirow}
\usepackage{siunitx}
\usepackage{url}
\usepackage{bm}
\usepackage{ctable} 
\usepackage{hyperref}
\usepackage{amsmath,amssymb,amsfonts}
\DeclareMathOperator*{\argmax}{arg\,max}
\usepackage{algorithm,algpseudocode}

\title{EvoSplit: An evolutionary approach to split a multi-label data set into disjoint subsets}

\author{
  Francisco Florez-Revuelta\\
  Department of Computing Technology\\
  University of Alicante, Spain\\
  \texttt{francisco.florez@ua.es} \\
}

\begin{document}
\maketitle

\begin{abstract}
This paper presents a new evolutionary approach, EvoSplit, for the distribution of multi-label data sets into disjoint subsets for supervised machine learning. Currently, data set providers either divide a data set randomly or using iterative stratification, a method that aims to maintain the label (or label pair) distribution of the original data set into the different subsets. Following the same aim, this paper first introduces a single-objective evolutionary approach that tries to obtain a split that maximizes the similarity between those distributions independently. Second, a new multi-objective evolutionary algorithm is presented to maximize the similarity considering simultaneously both distributions (labels and label pairs). Both approaches are validated using well-known multi-label data sets as well as large image data sets currently used in computer vision and machine learning applications. EvoSplit improves the splitting of a data set in comparison to the iterative stratification following different measures: Label Distribution, Label Pair Distribution, Examples Distribution, folds and fold-label pairs with zero positive examples. 
\end{abstract}

% keywords can be removed
\keywords {Multi-label data sets \and supervised learning \and machine learning \and evolutionary computation \and big data applications}

\section{Introduction}
\label{sec:introduction}

Supervised learning is the machine learning task of learning a function that maps an input to an output based on example input-output pairs~\cite{Russell2002}. An inducer receives a set of labeled examples as training data and makes predictions for unseen inputs~\cite{mohri2018,Kohavi1995}. Traditionally, each example is associated with a single label. However, in some problems an example might be associated with multiple labels. Multi-label machine learning has received significant attention in fields such as text categorization~\cite{Liu2017}, image classification~\cite{Wang_2016_CVPR}, health risk prediction~\cite{maxwell2017}, or electricity load monitoring~\cite{Tabatabaei2017}, among others. In computer vision in particular, there are more and more applications and available data sets that involve multi-label learning~\cite{COCO,wu2019tencent,Bustos2020}.

Formally~\cite{Zhang2014}, suppose $X=\mathbb{R}^d$ (or $\mathbb{Z}^d$) denotes the $d$-dimension instance space, and $Y = \{y_1, y_2,\dots, y_q\}$ denotes the label space with
$q$ possible class labels. The task of multi-label learning is to learn a function $h:X \rightarrow 2^Y$ from the multi-label training set $D = \{(x_i,Y_i) \vert 1 \leq i \leq m\}$. For each multi-label
example $(x_i,Y_i), x_i \in X$ is a $d$-dimensional feature vector
$(x_{i1}, x_{i2},\dots, x_{id})$ and $Y_i \subseteq Y$ is the set of labels associated with $x_i$. For any unseen instance $x \in X$, the multi-label
classifier $C(\cdot)$ predicts $C(x) \subseteq Y$ as the set of proper labels
for $x$.

In supervised learning, experiments typically involve a first step of distributing the examples of a data set into two or more disjoint subsets~\cite{sechidis2011stratification}. If a large amount of training data is available, the holdout method \cite{Kohavi1995} is used to distribute the examples into two mutually exclusive subsets called training set and test set, or holdout set. Sometimes the training set is also divided into two disjoint subsets to create a validation set. When training data is limited, $k$-fold cross-validation is used, which splits the data set into $k$ disjoint subsets of approximately equal size.

In single-label data sets, those disjoint subsets are built by distributing equally and randomly the examples in the original data set belonging to the different class labels. However, splitting a multi-label data set is not straightforward, as an over-represented class in one subset will be an under-represented in the other/s~\cite{Kohavi1995}.

Furthermore~\cite{sechidis2011stratification}, random distribution of multi-label training examples into subsets suffers from the following practical problem: it can lead to test subsets lacking even just one example of a rare label, which in turn causes calculation problems for a number of multi-label evaluation measures. The typical way these problems get by-passed in the literature is through complete removal of rare labels. This, however, implies that the performance of the learning systems on rare labels is unimportant, which is seldom true.

As mentioned by~\cite{szymanski2017network}, multi-label classification usually follows predetermined train/test splits set by data set providers, without the analysis in terms of how well the examples are distributed into those  train/test splits. Therefore, a method called stratification~\cite{Kohavi1995} or stratified sampling~\cite{sechidis2011stratification} was developed, in which a data set is split so that the proportion of examples of each class label in each subset is approximately equal to that in the complete data set. Stratification improves upon standard cross-validation both in terms of bias and variance, when compared to regular cross-validation~\cite{Kohavi1995}.

The data used for learning a classifier is often imbalanced~\cite{Tahir2012, Charte2015}. Thus, the class labels assigned to each instance are not equally represented. Traditionally, imbalanced classification has been faced through techniques such as resampling, cost-sensitive learning, and algorithmic-specific adaptations~\cite{Lopez2013, Liu2020}. In deep learning, data augmentation is a technique that has been successful in order to address imbalanced data sets~\cite{Shorten2019,Leng2017}.

Different measures have been proposed to estimate the degree of  multi-labelledness and the imbalance level of a data set~\cite{Zhang2014,Charte2013,Charte2018}. The \emph{label cardinality} $Card$ indicates the average number of labels per example~(Eq.~\eqref{eq:Card}). This measure can be normalized by the number $q$ of possible labels to obtain the~\emph{label density} $Dens$~(Eq.~\eqref{eq:Dens}). The~\emph{label diversity} $Div$ is the number of distinct label combinations that appear in the data set~(Eq.~\eqref{eq:Div}), which can also be normalized by the number of examples to indicate the proportion of distinct label sets~(Eq.~\eqref{eq:PDiv}). The \emph{Theoretical Complexity Score} $TCS$~(Eq.~\eqref{eq:TCS}) integrates the number of input features, the number of labels, and distinct label combinations into a single measure.

\begin{equation} 
\label{eq:Card}
Card(D)=\frac{\sum_{i=1}^{m} \left| Y_i \right|}{m}
\end{equation}

\begin{equation} 
\label{eq:Dens}
Dens(D)=\frac{Card(D)}{q}
\end{equation}

\begin{equation} 
\label{eq:Div}
Div(D)= \left| \{ Y^* \vert \exists x:(x,Y^*) \in D \textcolor{red}{\}}\right|
\end{equation}

\begin{equation} 
\label{eq:PDiv}
PDiv(D)= \frac{Div(D)}{m}
\end{equation}

\begin{equation} 
\label{eq:TCS}
TCS(D)= \log (m \times q \times Div(D))
\end{equation}

As mentioned in~\cite{Charte2015}, in binary classification the imbalance level is measured taking into account only two classes: the majority class and the minority class. In multi-label data sets the presence of the different labels can vary considerably. The \emph{average Imbalance Ratio} $avgIR$ is the average of the imbalance ratios ($IRLbl$) between the majority label and each label $Y_i \subseteq Y$~(Eq.~\eqref{eq:avgIR}). $IRLBl$ is equal to 1 for the most frequent label and greater for the other labels. Therefore, a larger value of the \emph{average Imbalance Ratio} represents a higher imbalance level in the data set.

\begin{equation} 
\label{eq:avgIR}
\begin{gathered}
avgIR(D)= \frac{\sum_{i=1}^{q}IRLbl(Y_i)}{q}, \\
IRLbl(y)= \frac{\argmax_{i=1}^{q} \left( \sum_{j=1}^{m}{h(Y_i,Y_j)} \right)} {\sum_{i=1}^{m} h(y,Y_i)}, \\
h(y,Y_k)=
    \begin{cases}
        1, & y = Y_k \\
        0, & y \neq Y_k
    \end{cases}
\end{gathered}
\end{equation}

The $SCUMBLE$ measure~(Eq.~\eqref{eq:SCUMBLE}) aims to quantify the imbalance variance among the labels present in each data sample. This measure allows the estimation of the level of co-occurrence between minority and majority labels, i.e.~if minority labels appear in their own or jointly with majority ones.

\begin{equation} 
\label{eq:SCUMBLE}
\begin{gathered}
SCUMBLE(D)= \frac{\sum_{i=1}^{m} SCUMBLE_i}{m}, \\
SCUMBLE_i= 1 - \frac{\left( \prod_{j=1}^{q} IRLbl(Y_j) \right) ^ {\frac{1}{q}}} {IRLbl(Y_i)} 
\end{gathered}
\end{equation}

\begin{figure}[t]
\centering
\includegraphics[width=0.5\columnwidth]{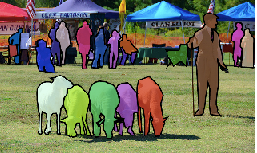}
\caption{Image from the Microsoft COCO data set in which there are several instances of multiple objects/class labels, e.g. sheep, people. (adapted from \url{https://cocodataset.org/\#detection-2020})}
\centering
\label{img:COCO}
\end{figure}

Table~\ref{tab:measures} presents these measures for well-known multi-label data sets, and recent large multi-label image data sets that are used in machine learning and computer vision applications. The complexity of these later data sets in terms of size, number of labels, cardinality, and diversity, is much higher than \emph{traditional} multi-label data sets. Some of these data sets, e.g Microsoft COCO, are labeled not only with the different classes that appear in one example but with the exact number of appearances of each class. This is the reason why the frequency of the label appearing the most in the Microsoft COCO data set is higher than one, as it appears several times, on average, per example~(Fig.~\ref{img:COCO}).

% Table generated by Excel2LaTeX from sheet 'Hoja1'
\begin{table}[t]
  \centering
  \caption{Imbalance measures of different multi-label data sets: size of data set (m), number of labels (q), max number of labels in an example of the data set (Max Labels), maximum frequency of a label in the data set (Max Frequency), label cardinality (Card), label density (Dens), label diversity (Div), normalized label diversity (PDiv), theoretical complexity score (TCS), average imbalance ratio per label (avgIR), and SCUMBLE. Well-known multi-label data sets~(top) have smaller size and lower complexity than data sets currently used in computer vision applications~(bottom). Data sets are ordered by theoretical complexity score. These data sets will be used in the validation of the different methods presented in this paper.}
  \setlength{\tabcolsep}{2pt}
  \footnotesize
\begin{tabular}{|l|c|c|c|c|c|c|c|c|c|c|c|}
\cline{2-12}    \multicolumn{1}{c|}{\multirow{2}[2]{*}{}} & \multirow{2}[2]{*}{\textbf{m}} & \multirow{2}[2]{*}{\textbf{q}} & \textbf{Max} & \textbf{Max} & \multirow{2}[2]{*}{\textbf{Card}} & \multirow{2}[2]{*}{\textbf{Dens}} & \multirow{2}[2]{*}{\textbf{Div}} & \multirow{2}[2]{*}{\textbf{PDiv}} & \multirow{2}[2]{*}{\textbf{TCS}} & \multirow{2}[2]{*}{\textbf{avgIR}} & \multirow{2}[2]{*}{\textbf{SCUMBLE}} \\
    \multicolumn{1}{c|}{} &       &       & \textbf{Labels} & \textbf{Frequency} &       &       &       &       &       &       &  \\
    \hline
    \textbf{emotions}~\cite{emotions} & \num{593}   & \num{6}      & \num{3}     & \num{0.45}  & \num{1.87}  & \num{0.311} & \num{27}    & \num{0.05}  &\num{9.7E+04} & \num{1.48}  & \num{0.01}\\
    \hline
    \textbf{scene}~\cite{scene} & \num{2407} &  $6$       & $3$     & $0.22$  & $1.07$  & \num{0.179} & $15$    & $0.01$  & \num{2.2E+05} & $1.25$  & $0.00$ \\
    \hline
    \textbf{genbase}~\cite{genbase} & $662$   & $27$     & $6$     & $0.26$  & $1.25$  & \num{0.046} & $32$    & $0.05$  & \num{5.7E+05} & $37.31$ & $0.03$ \\
    \hline
    \textbf{medical}~\cite{medical} & $978$   & $45$   & $3$     & $0.27$  & $1.25$  & \num{0.028} & $94$    & $0.10$  & \num{4.1E+06} & $89.50$ & $0.05$ \\
    \hline
    \textbf{yeast}~\cite{yeast} & \num{2417}  & $14$      & $11$    & $0.75$  & $4.24$  & \num{0.303} & $198$   & $0.08$  & \num{6.7E+06} & $7.20$  & $0.10$ \\
    \hline
    \textbf{enron}~\cite{enron} & \num{1702}  & $53$     & $12$    & $0.54$  & $3.38$  & \num{0.064} & $753$   & $0.44$  & \num{6.8E+07} & $73.95$ & $0.30$ \\
    \hline
    \textbf{rcv1subset4}~\cite{rcv1} & \num{6000}  & $101$    & $11$    & $0.25$  & $2.48$  & \num{0.025} & $816$   & $0.14$  & \num{4.9E+08} & $89.37$ & $0.22$ \\
    \hline
    \textbf{rcv1subset3}~\cite{rcv1} & \num{6000}  & $101$    & $12$    & $0.24$  & $2.61$  & \num{0.026} & $939$   & $0.16$  & \num{5.7E+08} & $68.33$ & $0.21$ \\
    \hline
    \textbf{rcv1subset5}~\cite{rcv1} & \num{6000}  & $101$    & $13$    & $0.25$  & $2.64$  & \num{0.026} & $946$   & $0.16$  & \num{5.7E+08} & $69.68$ & $0.24$ \\
    \hline
    \textbf{rcv1subset2}~\cite{rcv1} & \num{6000}  & $101$   & $12$    & $0.24$  & $2.63$  & \num{0.026} & $954$   & $0.16$  & \num{5.8E+08} & $45.51$ & $0.21$ \\
    \hline
    \textbf{rcv1subset1}~\cite{rcv1} & \num{6000}  & $101$    & $13$    & $0.23$  & $2.88$  & \num{0.029} & \num{1028}  & $0.17$  & \num{6.2E+08} & $54.49$ & $0.22$ \\
    \hline
    \textbf{tmc2007\_500}~\cite{tmc2007} & \num{28596} & $22$     & $10$    & $0.59$  & $2.22$  & \num{0.101} & \num{1172}  & $0.04$  & \num{7.4E+08} & $17.13$ & $0.19$ \\
    \hline
    \textbf{bibtex}~\cite{bibtex} & \num{7395}  & $159$   & $28$    & $0.14$  & $2.40$  & \num{0.015} & \num{2856}  & \num{0.39}  & \num{3.4E+09} & \num{12.50} & \num{0.09} \\
    \hline
    \textbf{Corel5k}~\cite{corel5k} & \num{5000}  & $374$   & $5$     & $0.22$  & $3.52$  & \num{0.009} & \num{3175}  & $0.64$  & \num{5.9E+09} & $189.57$ & $0.39$ \\
    \specialrule{.2em}{.0em}{.0em} 
\textbf{COCO}~\cite{COCO} & \num{123267} & $133$    & $98$    & $2.22$  & $11.25$ & \num{0.085} & \num{100676} & $0.82$  & \num{1.7E+12} & $75.33$ & $0.40$ \\
    \hline
    \textbf{Imagenet}~\cite{wu2019tencent,imagenet} &  \num{10756941} & \num{10592}  & $17$    & $0.91$  & $8.70$  & \num{0.001} & \num{10321} & \num{0.001} & \num{1.2E+15} & \num{52059.67} & $0.95$ \\
    \hline
    \textbf{OpenImages}~\cite{wu2019tencent,OpenImages} & \num{6941550} & \num{1345}  & $91$    & $0.50$  & $8.79$  & \num{0.007} & \num{885489} & $0.13$  & \num{8.3E+15} & \num{3015.02} & $0.42$ \\
    \hline
    \end{tabular}
  \label{tab:measures}%
\end{table}%

The remainder of this paper is organized as follows: Next, in Section~\ref{sec:related works} a review of available methods for the stratification of multi-label data sets is presented; Section~\ref{sec:EA} introduces and evaluates an evolutionary approach to obtain a stratified sampling of a data set; Section~\ref{sec:MOEA} proposes a multi-objective evolutionary algorithm to obtain an improved stratification; Section~\ref{sec:classification} evaluates the effect that this new splitting algorithm has on the classification metrics; and Section~\ref{sec:large_datasets} validates this latest evolutionary approach with large image data sets currently used in computer vision and machine learning applications, and compares the results with the official splits usually employed in the literature. Finally, Section~\ref{sec:conclusion} discusses the methods proposed in the paper and presents some future work.

\section{Related works}
\label{sec:related works}

The first approach to apply stratification to multi-label data sets was proposed by Sechidis et al.~\cite{sechidis2011stratification}. They proposed a greedy algorithm, Iterative Stratification (IS), that assigns iteratively examples from the original data set to each subset. The algorithm starts by taking the label with fewer examples in the data set. Those examples are, then, iteratively distributed among the different subsets based on the expected proportion of that label in each subset, i.e. an example is assigned to the subset that deviates more from the expected proportion. This process is repeated for each label until all the examples in the data set are distributed.

Sechidis et al. compared their approach against a random splitting using different measures. Following the notation presented in \cite{sechidis2011stratification}, let's consider a data set, $D$, annotated with a set of labels, $L = {\lambda_1,\dotsc,\lambda_q}$, a desired number $k$ of disjoints subsets $S_1,\dotsc,S_k$ of $D$, and a desired proportion of examples $r_1,\dotsc,r_k$ in each of these subsets. The desired number of examples at each subset $S_j$ is denoted as $c_j$ and is equal to $r_j\cdot\left|D\right|$ . The subsets of $D$ and $S_j$ that contain positive examples of label $\lambda_i$ are denoted $D_i$ and $S^i_j$ respectively.

The Label Distribution ($LD$) measure~\cite{sechidis2011stratification}~(Eq.~\eqref{eq1}) evaluates the extent to which the distribution of positive and negative examples of each label in each subset follows the distribution of that label in the whole data set. For each label $\lambda_i$, the measure computes the absolute difference between the ratio of positive to negative examples in each subset $S_j$ with the ratio of positive to negative examples in the whole data set $D$, and then averages the results across all labels. This measure is calculated as:

\begin{equation} 
\label{eq1}
LD=\frac{1}{q}\sum_{i=1}^{q}\left(\frac{1}{k} \sum_{j=1}^{k} \left| \frac{\left|S^i_j\right|}{\left|S_j\right|-\left|S^i_j\right|} -  \frac{\left|D^i\right|}{\left|D\right|-\left|D^i\right|}\right| \right)
\end{equation}

The Examples Distribution ($ED$) measure (Eq.~\eqref{eq4}) evaluates the extent to which the final number of examples in each subset $S_j$ deviates from the desired/expected number of examples in that subset.

\begin{equation} 
\label{eq4}
ED=\frac{1}{k}\sum_{j=1}^{k} \left| \left| S_j \right| - c_j \right|
\end{equation}

Other measures are the number of folds that contain at least one label with zero positive examples ($FZ$), and the number of fold-label pairs with zero positive examples ($FLZ$).

Using these measures, Sechidis et al. demonstrated that their iterative approach maintains better the ratio of positive to negative examples of each label in each subset ($LD$) and produces the smallest number of folds ($FZ$) and fold-label pairs ($FLZ$) with zero positive examples. However, their algorithm does not consider the desired number of examples in each subset as a hard constraint, getting worse results in terms of Examples Distribution ($ED$). They also mentioned that their approach might get worse results for multi-label classification methods that consider pairs of labels, e.g. Calibrated Label Ranking~\cite{furnkranz2008}, as their stratification method only considers the distribution of single labels.

Similarly to the measures presented in Section~\ref{sec:introduction}, measures could be defined not only for single labels appearing in a data set but also to higher order relations between them, i.e.~simultaneous appearance of labels (label pairs, triplets\dots), such as $Card^k$, $Dens^k$, $Div^k$, and $PDiv^k$. For instance, $Card^2(D)$ would indicate the average number of label pairs per example. Table~\ref{tab:measures2} shows these measures for order $2$ for the data sets previously analyzed. 

% Table generated by Excel2LaTeX from sheet 'Hoja1'
\begin{table}[t]
  \centering
  \caption{Measures of label pair imbalance of different multi-label data sets: label pair cardinality ($\text{Card}^2$), label pair density ($\text{Dens}^2$), maximum frequency of a label pair ($\text{Max Frequency}^2$), label pair diversity ($\text{Div}^2$) and normalized label pair diversity ($\text{PDiv}^2$)}
  \setlength{\tabcolsep}{3pt}
  \small
    \begin{tabular}{|l|c|c|c|c|c|}
\cline{2-6}    \multicolumn{1}{r|}{} & \multirow{2}[1]{*}{\textbf{$\text{Card}^2$}} & \multirow{2}[1]{*}{\textbf{$\text{Dens}^2$}} & \textbf{Max} & \multirow{2}[1]{*}{\textbf{$\text{Div}^2$}} & \multirow{2}[1]{*}{\textbf{$\text{PDiv}^2$}} \\
    \multicolumn{1}{r|}{} &       &       & \textbf{$\text{Frequency}^2$} &       &  \\
    \hline
    \textbf{emotions} & $1.04$  & \num{0.173} & $0.18$  & $14$    & $0.02$ \\
    \hline
    \textbf{scene} & $0.07$  & \num{0.012} & $0.03$  & $8$     & $0.00$ \\
    \hline
    \textbf{genbase} & $0.40$  & \num{0.015} & $0.04$  & $36$    & $0.05$ \\
    \hline
    \textbf{medical} & $0.26$  & \num{0.006} & $0.09$  & $63$    & $0.06$ \\
    \hline
    \textbf{yeast} & $8.09$  & \num{0.578} & $0.74$  & $89$    & $0.04$ \\
    \hline
    \textbf{enron} & $5.19$  & \num{0.098} & $0.34$  & $675$   & $0.40$ \\
    \hline
    \textbf{rcv1subset4} & $2.98$  & \num{0.029} & $0.09$  & \num{1452}  & $0.24$ \\
    \hline
    \textbf{rcv1subset3} & $3.27$  & \num{0.032} & $0.09$  & \num{1665}  & $0.28$ \\
    \hline
    \textbf{rcv1subset5} & $3.38$  & \num{0.033} & $0.09$  & \num{1680}  & $0.28$ \\
    \hline
    \textbf{rcv1subset2} & $3.34$  & \num{0.033} & $0.09$  & \num{1660}  & $0.28$ \\
    \hline
    \textbf{rcv1subset1} & $3.84$  & \num{0.038} & $0.08$  & \num{1802}  & $0.30$ \\
    \hline
    \textbf{tmc2007\_500} & $1.92$  & \num{0.087} & $0.16$  & $217$   & $0.01$ \\
    \hline
    \textbf{bibtex} & $3.11$  & \num{0.020} & $0.02$  & \num{4173}  & $0.56$ \\
    \hline
    \textbf{Corel5k} & $4.66$  & \num{0.012} & $0.05$  & \num{4342}  & $0.87$ \\
    \specialrule{.2em}{.0em}{.0em} 
    \textbf{COCO} & $26.69$ & \num{0.201} & $0.22$  & \num{8130}  & $0.07$ \\
    \hline
    \textbf{Imagenet} & $36.52$ & \num{0.003} & $0.88$  & \num{80587} & $0.01$ \\
    \hline
    \textbf{OpenImages} & $52.79$ & \num{0.039} & $0.45$  & \num{266470} & $0.04$ \\
    \hline
    \end{tabular}%
  \label{tab:measures2}%
\end{table}%

Given the limitation mentioned by Sechidis et al. regarding pairs of labels, Szymański and Kajdanowicz~\cite{szymanski2017network} extended the Iterative Stratification approach to take into account second-order relationships between labels, i.e.\ label pairs, and not just single labels into account when performing stratification. The proposed algorithm, Second Order Iterative Stratification (SOIS), behaves similarly to Sechidis et al.' stratification method but considering label pairs instead of single labels. The algorithm intends to maintain the same proportion of label pairs in each subset than in the original data set, considering at each point the label pair with fewer examples, distributing them among the subsets and repeating this process iteratively until no more label pairs are available. Finally, if there are examples with no label pairs these are assigned to the different subsets to comply with their expected size.

Szymański and Kajdanowicz compared SOIS with IS and random distribution using the same measures ($LD$, $ED$, $FZ$, $FLZ$). They also included a new measure, the Label Pair Distribution ($LPD$) (Eq.~\eqref{eq3}), an extension of the $LD$ measure that operates on positive and negative subsets of label pairs instead of labels. Given $E$ the set of label pairs appearing in the data set, $S^i_j$ and $D^i$ are the sets of samples that have the $i$-th label pair from $E$ assigned in subset $S_j$ and the entire data set respectively. In most cases, SOIS obtains better results than IS. 

\begin{equation} 
\label{eq3}
\small
LPD=\frac{1}{\left|E\right|}\sum_{i=1}^{\left|E\right|}\left(\frac{1}{k} \sum_{j=1}^{k} \left| \frac{\left|S^i_j\right|}{\left|S_j\right|-\left|S^i_j\right|} -  \frac{\left|D^i\right|}{\left|D\right|-\left|D^i\right|}\right| \right)
\end{equation}

\section{First approach: Single-objective evolutionary algorithm}
\label{sec:EA}

This work proposes an evolutionary algorithm (EA), EvoSplit, to obtain the distribution of a data set into disjoint subsets, considering their desired size as a hard constraint. The structure of the evolutionary algorithm follows the process presented in Algorithm~\ref{alg:alg1}. 

\begin{algorithm}
\caption{EvoSplit}
\begin{algorithmic}
\item[]
\State \textbf{Input:} Data set $D$, labels $L$, number of disjoint $k$ subsets, desired number $c$ of examples at each subset
\State \textbf{Output:} Disjoint subsets $S$
\item[]
\State \textbf{Initialize} the population with $n$ individuals generated randomly
\State \textbf{Rank} the population by fitness
\Repeat
	\State --------------------- Generate offspring --------------------		
		\State \textbf{Create} $c$ new individuals by crossover 				\State \textbf{Create} $m$ new individuals by mutation 				
		
		\State ------------------- Evaluate the offspring ------------------
		\State \textbf{Calculate} the fitness of the $c+m$ new individuals
	\State --------- Generate next generation's population ---------	
	\State \textbf{Rank} the $n+c+m$ individuals in the population by fitness
	\State \textbf{Select} the $n$ best individuals to generate a new population
\Until $generations\_without\_changes > gen_{max}$ 
\end{algorithmic}
\label{alg:alg1}
\end{algorithm}

\subsection{Characteristics of the algorithm}
Let $D$ be a multi-label data set, $k$ the desired number of disjoints subsets $S_1,\dotsc,S_k$ of $D$, and $c_1,\dotsc,c_k$ the desired number of examples at each subset. Each individual is encoded as an integer vector of size $|D|$, in which each gene represents the subset to which each example is assigned.

Different possibilities can be used to generate new individuals by crossover and mutation. EvoSplit selects parents by ranking, recombination is performed using 1-point crossover, and a mutation is carried out by reassigning randomly 1\% of the genes to a different subset. This process to generate new individuals would produce in most cases individuals that do not comply with the constraint of having $c_i$ examples in subset $S_i$, $i=1\dots k$. Therefore, a repairing process is applied to randomly reassign examples/genes to other subsets to fully comply with the constraint.

This work considers two different fitness functions: a variant of the Label Distribution ($LD$) and the Label Pair Distribution ($LPD$), which were introduced in Section~\ref{sec:related works}. The Label Distribution is appropriate for data sets in which a specific label can appear only once in an example. However, for data sets that might include in a particular example several instances of the same label, Eq.~\eqref{eq1} is not appropriate. This is the case, as it was shown earlier of well-known data sets in computer vision, as Microsoft COCO~\cite{COCO}. Therefore, the $LD$ has being modified to consider also data sets with this characteristic. 

Let's consider $\lambda^{S_j}_i$ and $\lambda^D_i$the number of appearances of label $\lambda_i$ in subset $S_j$ and data set $D$ respectively, $L^{S_j}$ and $L^D$ the total number of labels in subset $S_j$ and data set $D$ respectively. The modified Label Distribution measure, which is used as fitness function in EvoSplit, is then calculated following Eq.~\eqref{eq2}.   

\begin{equation} 
\label{eq2}
LD^\prime=\frac{1}{q}\sum_{i=1}^{q}\left(\frac{1}{k} \sum_{j=1}^{k} \left| \frac{\lambda^{S_j}_i}{L^{S_j}-\lambda^{S_j}_i} -  \frac{\lambda^D_i}{L^D-\lambda^D_i}\right|\right)
\end{equation}

We could proceed similarly with the $LPD$ measure. However, EvoSplit does not consider in its calculation the number of appearances of each label in each example, but only the co-occurrence of labels, as in the original $LPD$ measure. In this case, a variant of the $LPD$ would increase considerably the number of pair combinations and make difficult that different examples share the same pair. 

\subsection{Constraints}
\label{sec:constraints}
The application of evolutionary computation allows the introduction of constraints that all the individuals in the population must fulfill to be feasible. In Section~\ref{sec:introduction}, it was mentioned that the distribution of labels with few examples might lead to subsets lacking examples having that label, which can difficult validation and test of multi-label classifiers. Therefore, EvoSplit introduces an optional constraint to ensure that, if possible, all subsets contain, at least, one example of each label. For instance, if a data set has to be split into three subsets and a label only appears in three examples, each example will be distributed to a different subset. The constraint is not considered for those class labels which number of examples is lower than the number of subsets. Some other constraints could also be considered, if needed.

In case of generating an individual that does not fulfill the constraint, a repairing process similar to that explained before would be applied.

\subsection{Results}
\label{sec:EA_results}
Next, this work presents a comparison of the performance of the proposed evolutionary approach with other alternatives to split a data set into disjoint subsets, i.e.\ random and stratified (SOIS)\footnote{The stratified alternative has been obtained using the algorithm provided by scikit-multilearn \cite{Szymanski2019}.}. Similarly to the literature~\cite{sechidis2011stratification, szymanski2017network}, this work has evaluated the different methods considering 10-fold cross-validation of well-know multi-label data sets. 

For the evolutionary approaches, the parameters have been selected experimentally: 
\begin{itemize}
\item Size of the population ($n$): 50 
\item Individuals created by crossover ($c$): 10
\item Individuals created by mutation ($m$): 10
\item Number of generations without changes in the best individual ($gen_{max}$): 25
\end{itemize}

The splitting has been carried out using both the Label Distribution and the Label Pair Distribution as fitness functions. For each of these alternatives, results have been obtained without and with the constraint presented in Section~\ref{sec:constraints} that tries to ensure that, if possible, all folds have examples for all the labels. For each alternative, given the probabilistic behavior of evolutionary algorithms, the best result of five runs of the algorithm has been selected.

Table~\ref{tab:EA_cross_validation_LD} shows the results in the case of using the Label Distribution as fitness function. In this case, the optimization algorithm tries to create subsets in which the proportion of examples of each class is close to that in the complete data set. In both cases, whether considering or not the constraint, the evolutionary approach obtains better results than any other method. Something similar happens~((Table~\ref{tab:EA_cross_validation_LPD})) when the Label Pair Distribution is employed as fitness function, i.e. the algorithm tries to approximate the propotion of label pairs. However, in most cases, when the evolutionary algorithm tries to improve the distribution of single labels (by using $LD$) fails to distribute label pairs better than the Stratification method. Something similar happens when the fitness function is the $LPD$ measure and the results are measured in terms of $LD$.  

% Table generated by Excel2LaTeX from sheet 'Tablas finales'
\begin{table}[t]
  \centering
  \setlength{\tabcolsep}{4pt}
  \footnotesize
  \caption{Label Distribution of different splitting algorithms. The Stratification method is considered as the baseline. In bold the results that are better than the Stratification method. In all cases, the evolutionary algorithm that uses $LD$ as fitness function obtains the best results. This does not happen when the $LPD$ is used as fitness function.}
    \begin{tabular}{|l|r|r|r|r|r|r|}
\cline{2-7} \multicolumn{1}{r|}{} & \multicolumn{1}{c|}{\multirow{3}[6]{*}{\textbf{Random}}} & \multicolumn{1}{c|}{\multirow{3}[6]{*}{\textbf{Stratification}}} & \multicolumn{4}{c|}{\textbf{Evolutionary algorithm}} \\
\cline{4-7} \multicolumn{1}{r|}{} & & & \multicolumn{2}{c|}{\textbf{Label Distribution}} & \multicolumn{2}{c|}{\textbf{Label Pair Distribution}} \\
\cline{4-7} \multicolumn{1}{r|}{} & & & \multicolumn{1}{c|}{\textbf{unconstrained}} & \multicolumn{1}{c|}{\textbf{constrained}} & \multicolumn{1}{c|}{\textbf{unconstrained}} & \multicolumn{1}{c|}{\textbf{constrained}} \\
\hline
\textbf{emotions} & $3.77 \times 10^{-2}$ & $6.45 \times 10^{-3}$ & \bm{$2.87 \times 10^{-3}$} & \bm{$2.85 \times 10^{-3}$} & $1.84 \times 10^{-2}$ & $1.79 \times 10^{-2}$ \\
\hline
\textbf{scene} & $2.68 \times 10^{-2}$ & $1.69 \times 10^{-3}$ & \bm{$1.27 \times 10^{-3}$} & \bm{$1.42 \times 10^{-3}$} & $2.46 \times 10^{-2}$ & $2.38 \times 10^{-2}$ \\
\hline
\textbf{genbase} & $1.38 \times 10^{-2}$ & $4.75 \times 10^{-3}$ & \bm{$4.54 \times 10^{-3}$} & \bm{$4.48 \times 10^{-3}$} & $1.03 \times 10^{-2}$ & $1.30 \times 10^{-2}$ \\
\hline
\textbf{medical} & $7.68 \times 10^{-3}$ & $2.96 \times 10^{-3}$ & \bm{$2.88 \times 10^{-3}$} & \bm{$2.79 \times 10^{-3}$} & $7.56 \times 10^{-3}$ & $7.61 \times 10^{-3}$ \\
\hline
\textbf{yeast} & $5.72 \times 10^{-3}$ & $1.47 \times 10^{-3}$ & \bm{$5.24 \times 10^{-4}$} & \bm{$5.12 \times 10^{-4}$} & \bm{$1.35 \times 10^{-3}$} & \bm{$1.29 \times 10^{-3}$} \\
\hline
\textbf{enron} & $2.89 \times 10^{-3}$ & $1.79 \times 10^{-3}$ & \bm{$8.90 \times 10^{-4}$} & \bm{$8.34 \times 10^{-4}$} & \bm{$1.46 \times 10^{-3}$} & \bm{$1.45 \times 10^{-3}$} \\
\hline
\textbf{rcv1subset4} & $1.55 \times 10^{-3}$ & $4.24 \times 10^{-4}$ & \bm{$3.72 \times 10^{-4}$} & \bm{$3.30 \times 10^{-4}$} & $5.87 \times 10^{-4}$ & $5.73 \times 10^{-4}$ \\
\hline
\textbf{rcv1subset3} & $1.60 \times 10^{-3}$ & $4.96 \times 10^{-4}$ & \bm{$3.42 \times 10^{-4}$} & \bm{$3.30 \times 10^{-4}$} & $5.84 \times 10^{-4}$ & $5.41 \times 10^{-4}$ \\
\hline
\textbf{rcv1subset5} & $1.55 \times 10^{-3}$ & $4.21 \times 10^{-4}$ & \bm{$3.20 \times 10^{-4}$} & \bm{$3.14 \times 10^{-4}$} & $5.14 \times 10^{-4}$ & $4.95 \times 10^{-4}$ \\
\hline
\textbf{rcv1subset2} & $1.53 \times 10^{-3}$ & $4.07 \times 10^{-4}$ & \bm{$3.39 \times 10^{-4}$} & \bm{$3.21 \times 10^{-4}$} & $5.30 \times 10^{-4}$ & $5.05 \times 10^{-4}$ \\
\hline
\textbf{rcv1subset1} & $1.67 \times 10^{-3}$ & $4.32 \times 10^{-4}$ & \bm{$3.26 \times 10^{-4}$} & \bm{$2.95 \times 10^{-4}$} & $5.04 \times 10^{-4}$ & $4.59 \times 10^{-4}$ \\
\hline
\textbf{tmc2007\_500} & $1.76 \times 10^{-3}$ & $2.23 \times 10^{-4}$ & \bm{$1.74 \times 10^{-4}$} & \bm{$1.61 \times 10^{-4}$} & $8.05 \times 10^{-4}$ & $7.43 \times 10^{-4}$ \\
\hline
\textbf{bibtex} & $1.35 \times 10^{-3}$ & $3.62 \times 10^{-4}$ & \bm{$2.40 \times 10^{-4}$} & \bm{$2.39 \times 10^{-4}$} & $7.81 \times 10^{-4}$ & $7.94 \times 10^{-4}$ \\
\hline
\textbf{Corel5k} & $6.36 \times 10^{-4}$ & $4.39 \times 10^{-4}$ & \bm{$2.68 \times 10^{-4}$} & \bm{$2.48 \times 10^{-4}$} & \bm{$3.71 \times 10^{-4}$} & \bm{$3.44 \times 10^{-4}$} \\
\hline
\end{tabular}%
\label{tab:EA_cross_validation_LD}%
\end{table}%

% Table generated by Excel2LaTeX from sheet 'Tablas finales'
\begin{table}[t]
  \centering
  \setlength{\tabcolsep}{4pt}
  \footnotesize
    \caption{Label Pair Distribution of different splitting algorithms.  In bold the results that are better than the stratification method. In all cases, the evolutionary algorithm that uses $LPD$ as fitness function obtains the best results. This does not happen when the $LD$ is used as fitness function.}
\begin{tabular}{|l|r|r|r|r|r|r|}
\cline{2-7} \multicolumn{1}{r|}{} & \multicolumn{1}{c|}{\multirow{3}[6]{*}{\textbf{Random}}} & \multicolumn{1}{c|}{\multirow{3}[6]{*}{\textbf{Stratification}}} & \multicolumn{4}{c|}{\textbf{Evolutionary algorithm}} \\
\cline{4-7} \multicolumn{1}{r|}{} & & & \multicolumn{2}{c|}{\textbf{Label Distribution}} & \multicolumn{2}{c|}{\textbf{Label Pair Distribution}} \\
\cline{4-7} \multicolumn{1}{r|}{} & & & \multicolumn{1}{c|}{\textbf{unconstrained}} & \multicolumn{1}{c|}{\textbf{constrained}} & \multicolumn{1}{c|}{\textbf{unconstrained}} & \multicolumn{1}{c|}{\textbf{constrained}} \\
\hline
\textbf{emotions} & $2.64 \times 10^{-2}$ & $6.61 \times 10^{-3}$ & $1.54 \times 10^{-2}$ & $1.93 \times 10^{-2}$ & \bm{$4.81 \times 10^{-3}$} & \bm{$4.63 \times 10^{-3}$} \\
\hline
\textbf{scene} & $1.08 \times 10^{-1}$ & $2.83 \times 10^{-2}$ & $7.84 \times 10^{-2}$ & $8.11 \times 10^{-2}$ & \bm{$1.97 \times 10^{-2}$} & \bm{$2.11 \times 10^{-2}$} \\
\hline
\textbf{genbase} & $1.98 \times 10^{-2}$ & $1.73 \times 10^{-2}$ & \bm{$1.71 \times 10^{-2}$} & $1.74 \times 10^{-2}$ & \bm{$1.37 \times 10^{-2}$} & \bm{$1.37 \times 10^{-2}$} \\
\hline
\textbf{medical} & $1.61 \times 10^{-2}$ & $1.28 \times 10^{-2}$ & $1.53 \times 10^{-2}$ & $1.46 \times 10^{-2}$ & \bm{$1.17 \times 10^{-2}$} & \bm{$1.17 \times 10^{-2}$} \\
\hline
\textbf{yeast} & $1.57 \times 10^{-3}$ & $6.45 \times 10^{-4}$ & $8.39 \times 10^{-4}$ & $8.81 \times 10^{-4}$ & \bm{$3.60 \times 10^{-4}$} & \bm{$3.35 \times 10^{-4}$} \\
\hline
\textbf{enron} & $6.68 \times 10^{-4}$ & $5.61 \times 10^{-4}$ & $5.66 \times 10^{-4}$ & \bm{$5.57 \times 10^{-4}$} & \bm{$4.16 \times 10^{-4}$} & \bm{$4.19 \times 10^{-4}$} \\
\hline
\textbf{rcv1subset4} & $3.35 \times 10^{-4}$ & $2.10 \times 10^{-4}$ & $2.63 \times 10^{-4}$ & $2.54 \times 10^{-4}$ & \bm{$1.87 \times 10^{-4}$} & \bm{$1.86 \times 10^{-4}$} \\
\hline
\textbf{rcv1subset3} & $2.98 \times 10^{-4}$ & $2.29 \times 10^{-4}$ & $2.40 \times 10^{-4}$ & $2.41 \times 10^{-4}$ & \bm{$1.73 \times 10^{-4}$} & \bm{$1.72 \times 10^{-4}$} \\
\hline
\textbf{rcv1subset5} & $2.85 \times 10^{-4}$ & $1.96 \times 10^{-4}$ & $2.34 \times 10^{-4}$ & $2.30 \times 10^{-4}$ & \bm{$1.69 \times 10^{-4}$} & \bm{$1.69 \times 10^{-4}$} \\
\hline
\textbf{rcv1subset2} & $2.92 \times 10^{-4}$ & $1.85 \times 10^{-4}$ & $2.40 \times 10^{-4}$ & $2.38 \times 10^{-4}$ & \bm{$1.73 \times 10^{-4}$} & \bm{$1.70 \times 10^{-4}$} \\
\hline
\textbf{rcv1subset1} & $2.70 \times 10^{-4}$ & $1.78 \times 10^{-4}$ & $2.11 \times 10^{-4}$ & $2.05 \times 10^{-4}$ & \bm{$1.51 \times 10^{-4}$} & \bm{$1.52 \times 10^{-4}$} \\
\hline
\textbf{tmc2007\_500} & $4.69 \times 10^{-4}$ & $1.94 \times 10^{-4}$ & $3.82 \times 10^{-4}$ & $3.71 \times 10^{-4}$ & \bm{$1.30 \times 10^{-4}$} & \bm{$1.18 \times 10^{-4}$} \\
\hline
\textbf{bibtex} & $1.96 \times 10^{-4}$ & $1.68 \times 10^{-4}$ & $1.82 \times 10^{-4}$ & $1.80 \times 10^{-4}$ & \bm{$1.44 \times 10^{-4}$} & \bm{$1.45 \times 10^{-4}$} \\
\hline
\textbf{Corel5k} & $1.72 \times 10^{-4}$ & $1.43 \times 10^{-4}$ & $1.59 \times 10^{-4}$ & $1.59 \times 10^{-4}$ & \bm{$1.29 \times 10^{-4}$} & \bm{$1.29 \times 10^{-4}$} \\
\hline
\end{tabular}%
\label{tab:EA_cross_validation_LPD}%
\end{table}%

It is worth mentioning that the Stratification method does not consider the desired number of samples per fold as a hard constraint. Therefore, the final sizes of the subsets might deviate from the pre-established ones, as measured by the Examples Distribution and shown in Table~\ref{tab:ED_cross_validation}. 

% Table generated by Excel2LaTeX from sheet 'Tablas finales'
\begin{table}[t]
  \centering
  \caption{Examples Distribution of different splitting algorithms. Only the stratification method deviates from zero.}
        \begin{tabular}{|l|c|c|c|}
\cline{2-4} \multicolumn{1}{l|}{} & \textbf{Random} & \textbf{Stratification} & \textbf{All EAs} \\
\hline
\textbf{emotions} & 0 & 1.2 & 0 \\
\hline
\textbf{scene} & 0 & 1.4 & 0 \\
\hline
\textbf{genbase} & 0 & 0.8 & 0 \\
\hline
\textbf{medical} & 0 & 1.2 & 0 \\
\hline
\textbf{yeast} & 0 & 4.2 & 0 \\
\hline
\textbf{enron} & 0 & 3.8 & 0 \\
\hline
\textbf{rcv1subset4} & 0 & 3.8 & 0 \\
\hline
\textbf{rcv1subset3} & 0 & 4.8 & 0 \\
\hline
\textbf{rcv1subset5} & 0 & 6.6 & 0 \\
\hline
\textbf{rcv1subset2} & 0 & 3.6 & 0 \\
\hline
\textbf{rcv1subset1} & 0 & 7.2 & 0 \\
\hline
\textbf{tmc2007\_500} & 0 & 19.8 & 0 \\
\hline
\textbf{bibtex} & 0 & 14 & 0 \\
\hline
\textbf{Corel5k} & 0 & 6.2 & 0 \\
\hline
\end{tabular}%
\label{tab:ED_cross_validation}%
\end{table}%

Following \cite{sechidis2011stratification}, besides using the Label Distribution and the Label Pair Distribution, the result of each alternative is also measured (see Tables~\ref{tab:FZ_cross_validation}~and~\ref{tab:FLZ_cross_validation}) in terms of the number of folds that contain at least one label with zero positive examples ($FZ$), and the number of fold-label pairs with zero positive examples ($FLZ$). The $FLZ$ measure shows the effect of introducing the constraint of ensuring that subsets contain, if possible, one example of each label. In the constrained version of EvoSplit, the $FLZ$ values are lower than with all the other methods.

% Table generated by Excel2LaTeX from sheet 'Tablas finales'
\begin{table}[t]
  \centering
  \caption{Number of folds that contain at least one label with zero positive examples ($FZ$). All the splitting algorithms obtain the same results.}
        \begin{tabular}{|l|c|}
\cline{2-2} \multicolumn{1}{c|}{} & \multicolumn{1}{l|}{\textbf{All cases}} \\
\hline
\textbf{emotions} & 0 \\
\hline
\textbf{scene} & 0 \\
\hline
\textbf{genbase} & 10 \\
\hline
\textbf{medical} & 10 \\
\hline
\textbf{yeast} & 0 \\
\hline
\textbf{enron} & 10 \\
\hline
\textbf{rcv1subset4} & 10 \\
\hline
\textbf{rcv1subset3} & 10 \\
\hline
\textbf{rcv1subset5} & 10 \\
\hline
\textbf{rcv1subset2} & 10 \\
\hline
\textbf{rcv1subset1} & 10 \\
\hline
\textbf{tmc2007\_500} & 0 \\
\hline
\textbf{bibtex} & 0 \\
\hline
\textbf{Corel5k} & 10 \\
\hline
\end{tabular}%

\label{tab:FZ_cross_validation}%
\end{table}%

% Table generated by Excel2LaTeX from sheet 'Tablas finales'
\begin{table}[t]
  \centering
  \setlength{\tabcolsep}{4pt}
  \footnotesize
  \caption{Number of fold-label pairs with zero positive examples ($FLZ$) of different splitting algorithms. In bold the results that are better than the stratification method. Between brackets those results that are worse.}
    \begin{tabular}{|l|c|c|c|c|c|c|}
\cline{2-7} \multicolumn{1}{r|}{} & \multirow{3}[6]{*}{\textbf{Random}} & \multirow{3}[6]{*}{\textbf{Stratification}} & \multicolumn{4}{c|}{\textbf{Evolutionary algorithm}} \\
\cline{4-7} \multicolumn{1}{r|}{} & & & \multicolumn{2}{c|}{\textbf{Label Distribution}} & \multicolumn{2}{c|}{\textbf{Label Pair Distribution}} \\
\cline{4-7} \multicolumn{1}{r|}{} & & & \multicolumn{1}{c|}{\textbf{unconstrained}} & \multicolumn{1}{c|}{\textbf{constrained}} & \multicolumn{1}{c|}{\textbf{unconstrained}} & \multicolumn{1}{c|}{\textbf{constrained}} \\
\hline
\textbf{emotions} & 0 & 0 & 0 & 0 & 0 & 0 \\
\hline
\textbf{scene} & 0 & 0 & 0 & 0 & 0 & 0 \\
\hline
\textbf{genbase} & 85 & 76 & \textbf{74} & \textbf{75} & \textbf{74} & \textbf{73} \\
\hline
\textbf{medical} & 202 & 174 & (176) & (175) & (200) & (180) \\
\hline
\textbf{yeast} & 0 & 0 & 0 & 0 & 0 & 0 \\
\hline
\textbf{enron} & 87 & 77 & \textbf{59} & \textbf{53} & \textbf{62} & \textbf{55} \\
\hline
\textbf{rcv1subset4} & 106 & 68 & (78) & \textbf{63} & \textbf{66} & \textbf{62} \\
\hline
\textbf{rcv1subset3} & 77 & 57 & \textbf{54} & \textbf{40} & \textbf{44} & \textbf{39} \\
\hline
\textbf{rcv1subset5} & 83 & 50 & \textbf{49} & \textbf{44} & \textbf{48} & \textbf{47} \\
\hline
\textbf{rcv1subset2} & 69 & 35 & (36) & \textbf{30} & \textbf{34} & \textbf{33} \\
\hline
\textbf{rcv1subset1} & 74 & 59 & \textbf{56} & \textbf{43} & \textbf{50} & \textbf{44} \\
\hline
\textbf{tmc2007\_500} & 0 & 0 & 0 & 0 & 0 & 0 \\
\hline
\textbf{bibtex} & 2 & 0 & 0 & 0 & 0 & 0 \\
\hline
\textbf{Corel5k} & 1157 & 1019 & \textbf{925} & \textbf{843} & \textbf{948} & \textbf{866} \\
\hline
\end{tabular}%
\label{tab:FLZ_cross_validation}%
\end{table}%

Tables~\ref{tab:EA_cross_validation_LD}~and~\ref{tab:EA_cross_validation_LPD} show that the evolutionary approach gets better results than Stratification only for the metric that is employed as fitness function. There are only some few cases in which good results are obtained for both metrics, the Label Distribution and the Label Pair Distribution simultaneously. On the other hand, the Stratification method obtains splits that behave well for both metrics, even if the best results for each metric are obtained with the evolutionary approaches. This is probably due to the process that the Stratification method follows to distribute examples to subsets: first, considering label pairs in the distribution and, later, assigning remaining examples based on single labels, i.e. taking into consideration both labels and label pairs in the splitting.

\section{Second approach: Multi-Objective Evolutionary Algorithm}
\label{sec:MOEA}

Then, it seems appropriate to consider both statistical measures, the Label Distribution and the Label Pair Distribution, in the optimization algorithm to split a data set into disjoint subsets. Multi-objective optimization problems are those problems where the goal is to optimize simultaneously several objective functions. These different functions have conflicting objectives, i.e. optimizing one affects the others. Therefore, there is not a unique solution but a set of solutions. The set of solutions in which the different objective components cannot be simultaneously improved constitute a Pareto front. Each solution in the Pareto front represents a trade-off between the different objectives. Similarly to evolutionary algorithms for single objective problems, multi-objective evolutionary algorithms (MOEA)~\cite{coello2007} are heuristic algorithms to solve problems with multiple objective functions. The three goals of an
MOEA are~\cite{Trivedi2017}: 1) to find a set of solutions as close as possible to the Pareto front (known as convergence); 2) to maintain a diverse population that contains dissimilar individuals to promote exploration and to avoid poor performance due to premature convergence (known as diversity); and 3) to obtain a set of solutions that spreads in a more uniform way over the Pareto front (known as coverage). Several MOEAs have been proposed in the literature. This work employs the Non-dominated Sorting Genetic Algorithm II~ (NSGA-II)~\cite{Deb2020}.

NSGA-II has the three following features: 1) it uses an elitist principle, i.e. the elites of a population are given the opportunity to be carried to the next generation; 2) it uses an explicit diversity preserving mechanism (Crowding distance); and 3) it emphasizes the non-dominated solutions.

\begin{figure}[b]
\centering
\includegraphics[width=0.5\columnwidth]{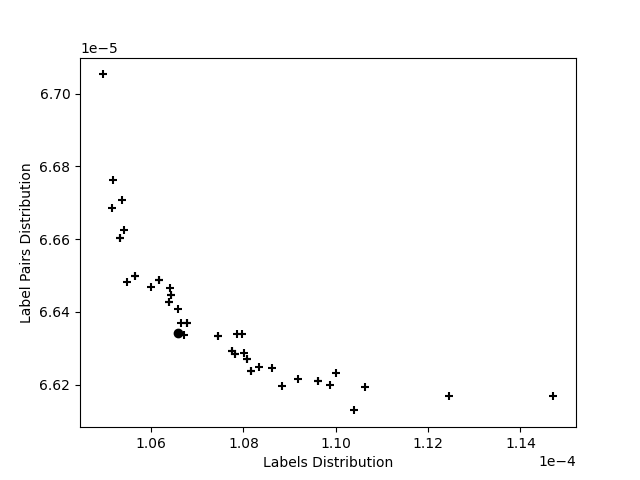}
\caption{Final set of solutions for a run of the MOEA algorithm to split the Corel5k data set. Selected global optimum is marked with a circle.}
\centering
\label{img:MOEA_pareto}
\end{figure}

Therefore, this work employs NSGA-II to distribute the data set into subsets optimizing simultaneously both the Label Distribution and the Label Pair Distribution. The algorithm will obtain a set of solutions, some of them optimizing one over the other objective and vice versa. From these set of solutions, EvoSplit selects the solution closer (using Euclidean distance) to the coordinates origin~(Fig.~\ref{img:MOEA_pareto}).

This work has employed the implementation of NSGA-II offered by  pymoo~\cite{pymoo}, a multi-objective optimization framework in Python, using the same parameters in terms of individuals, size of offspring and ending condition presented in Section~\ref{sec:EA_results}. 

\subsection{Results}

% Table generated by Excel2LaTeX from sheet 'Tablas finales'
\begin{table}[t]
  \centering
  \caption{Evaluation of the MOEA approach without considering the constraint. In bold the results that are better than the Stratification method.}
    \begin{tabular}{|l|r|r|c|c|}
\cline{2-5} \multicolumn{1}{r|}{} & \multicolumn{1}{c|}{\textbf{LD}} & \multicolumn{1}{c|}{\textbf{LPD}} & \textbf{FZ} & \textbf{FLZ} \\
\hline
\textbf{emotions} & \bm{$5.28 \times 10^{-3}$} & $7.43 \times 10^{-3}$ & 0 & \textbf{0} \\
\hline
\textbf{scene} & \bm{$1.77 \times 10^{-3}$} & \bm{$2.18 \times 10^{-2}$} & 0 & \textbf{0} \\
\hline
\textbf{genbase} & \bm{$4.70 \times 10^{-3}$} & \bm{$1.42 \times 10^{-2}$} & 10 & \textbf{74} \\
\hline
\textbf{medical} & $3.39 \times 10^{-3}$ & \bm{$1.19 \times 10^{-2}$} & 10 & 179 \\
\hline
\textbf{yeast} & \bm{$6.56 \times 10^{-4}$} & \bm{$4.80 \times 10^{-4}$} & 0 & \textbf{0} \\
\hline
\textbf{enron} & \bm{$8.28 \times 10^{-4}$} & \bm{$4.60 \times 10^{-4}$} & 10 & \textbf{58} \\
\hline
\textbf{rcv1subset4} & \bm{$2.93 \times 10^{-4}$} & \bm{$2.04 \times 10^{-4}$} & 10 & 71 \\
\hline
\textbf{rcv1subset3} & \bm{$2.93 \times 10^{-4}$} & \bm{$1.87 \times 10^{-4}$} & 10 & \textbf{42} \\
\hline
\textbf{rcv1subset5} & \bm{$2.70 \times 10^{-4}$} & \bm{$1.81 \times 10^{-4}$} & 10 & \textbf{47} \\
\hline
\textbf{rcv1subset2} & \bm{$2.96 \times 10^{-4}$} & $1.89 \times 10^{-4}$ & 10 & 36 \\
\hline
\textbf{rcv1subset1} & \bm{$2.94 \times 10^{-4}$} & \bm{$1.68 \times 10^{-4}$} & 10 & \textbf{50} \\
\hline
\textbf{tmc2007\_500} & \bm{$1.27 \times 10^{-4}$} & \bm{$1.89 \times 10^{-4}$} & 0 & \textbf{0} \\
\hline
\textbf{bibtex} & \bm{$2.68 \times 10^{-4}$} & \bm{$1.53 \times 10^{-4}$} & 0 & \textbf{0} \\
\hline
\textbf{Corel5k} & \bm{$2.84 \times 10^{-4}$} & \bm{$1.41 \times 10^{-4}$} & 10 & \textbf{907} \\
\hline
\end{tabular}%
\label{tab:MOEA_cross_validation_unconstrained}%
\end{table}%

% Table generated by Excel2LaTeX from sheet 'Tablas finales'
\begin{table}[t]
  \centering
  \caption{Evaluation of the MOEA approach considering the constraint. In bold the results that are better than the Stratification method.}
    \begin{tabular}{|l|r|r|c|c|}
\cline{2-5} \multicolumn{1}{r|}{} & \multicolumn{1}{c|}{\textbf{LD}} & \multicolumn{1}{c|}{\textbf{LPD}} & \textbf{FZ} & \textbf{FLZ} \\
\hline
\textbf{genbase} & $5.12 \times 10^{-3}$ & \bm{$1.40 \times 10^{-2}$} & 10 & \textbf{73} \\
\hline
\textbf{medical} & \bm{$2.96 \times 10^{-3}$} & \bm{$1.19 \times 10^{-2}$} & 10 & 175 \\
\hline
\textbf{enron} & \bm{$8.29 \times 10^{-4}$} & \bm{$4.63 \times 10^{-4}$} & 10 & \textbf{53} \\
\hline
\textbf{rcv1subset4} & \bm{$3.11 \times 10^{-4}$} & \bm{$2.06 \times 10^{-4}$} & 10 & \textbf{61} \\
\hline
\textbf{rcv1subset3} & \bm{$2.70 \times 10^{-4}$} & \bm{$1.85 \times 10^{-4}$} & 10 & \textbf{38} \\
\hline
\textbf{rcv1subset5} & \bm{$2.69 \times 10^{-4}$} & \bm{$1.84 \times 10^{-4}$} & 10 & \textbf{45} \\
\hline
\textbf{rcv1subset2} & \bm{$2.87 \times 10^{-4}$} & $1.89 \times 10^{-4}$ & 10 & \textbf{31} \\
\hline
\textbf{rcv1subset1} & \bm{$2.54 \times 10^{-4}$} & \bm{$1.63 \times 10^{-4}$} & 10 & \textbf{41} \\
\hline
\textbf{Corel5k} & \bm{$2.52 \times 10^{-4}$} & \bm{$1.39 \times 10^{-4}$} & 10 & \textbf{841} \\
\hline
\end{tabular}%

\label{tab:MOEA_cross_validation_constrained}%
\end{table}%

Table~\ref{tab:MOEA_cross_validation_unconstrained} shows the results obtained for the different measures of the splits obtained using the MOEA unconstrained approach. The Examples Distribution measure is not shown as it is always zero, as with the previous evolutionary approaches. The obtained results are, in most cases, better that those obtained with the Stratification method. In this approach, unlike the previous single-objective evolutionary alternatives, results are good in terms of both $LD$ and $LPD$. The MOEA approach obtains results in terms of $LD$ close to those obtained by the single-objective approach optimizing only $LD$ (see Table~\ref{tab:EA_cross_validation_LD}), and close in terms of $LPD$ when the optimization is based only in $LPD$ (see Table~\ref{tab:EA_cross_validation_LPD}). These are more balanced results than those obtained with the single-objective evolutionary approaches, i.e. a good result in one of the measures does not affect a good result in the other one. Additionally, results are also quite similar in terms of $FZ$ and $FLZ$. For those data sets with $FLZ$ different to zero, Table~\ref{tab:MOEA_cross_validation_constrained} shows the results obtained with the constrained alternative. For some data sets, $FLZ$ is reduced without affecting considerably the $LD$ and $LPD$ measures.

\section{Evaluation of classification}
\label{sec:classification}

Next, this work evaluates the effect that the distribution of examples in a multi-label data set into subsets has on the classification metrics. Similarly to~\cite{szymanski2017network}, the evaluation is carried out employing two standard multi-label classification algorithms: Binary Relevance~(BR) and Label Powerset~(LP). The results of the classification are measured using recall, precision, $F_1$ score and Jaccard index. These metrics are calculated using both micro-average (i.e. calculate metrics globally by counting the total true positives, false negatives and false positives) and macro-average (i.e. calculate metrics for each label, and then their unweighted mean). The variance in the classification across different folds is also analyzed, which provides information about the generalization stability of the different approaches. 

\begin{figure}[t]
\centering
\includegraphics[width=\columnwidth]{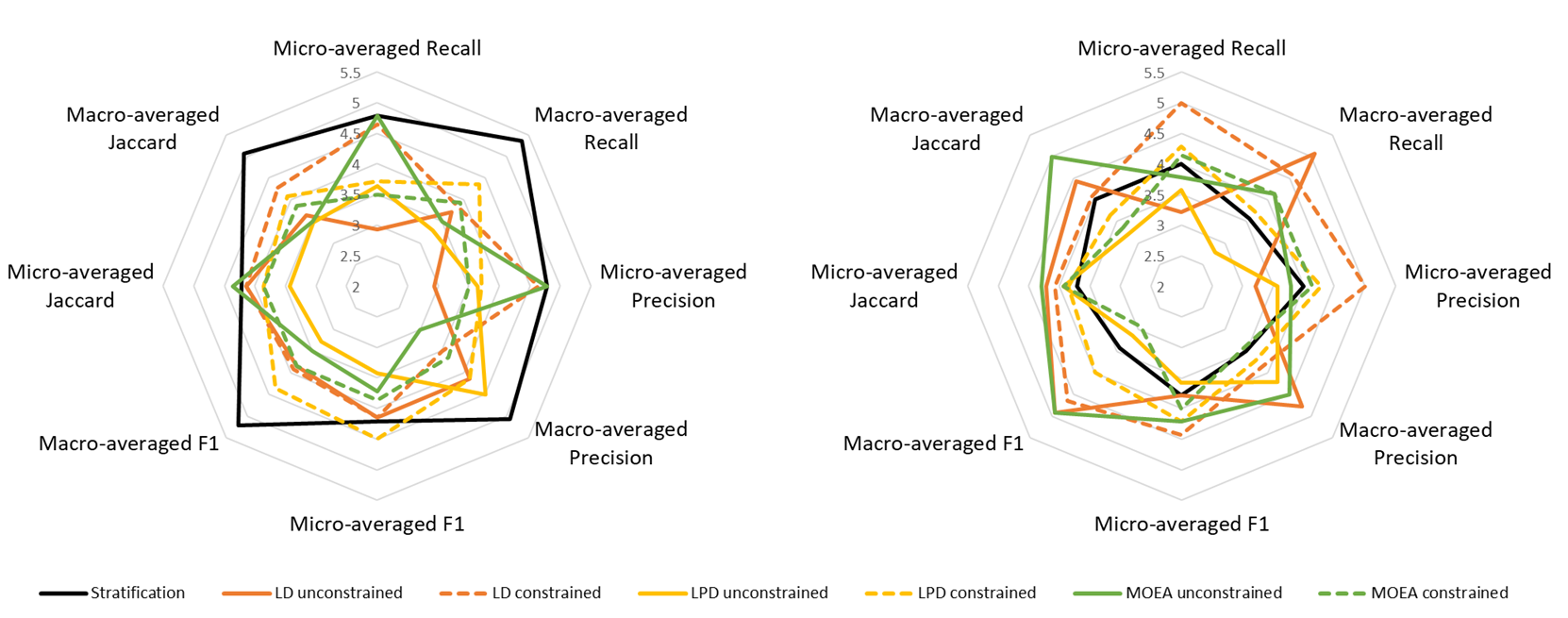}
\caption{Average ranks of the different methods when classification is performed using Binary Relevance. Left figure shows the ranks regarding classification metrics. Right figure shows the ranks regarding the variance over folds.}
\label{img:BinaryRelevance}
\end{figure}

\begin{figure}[t]
\centering
\includegraphics[width=\columnwidth]{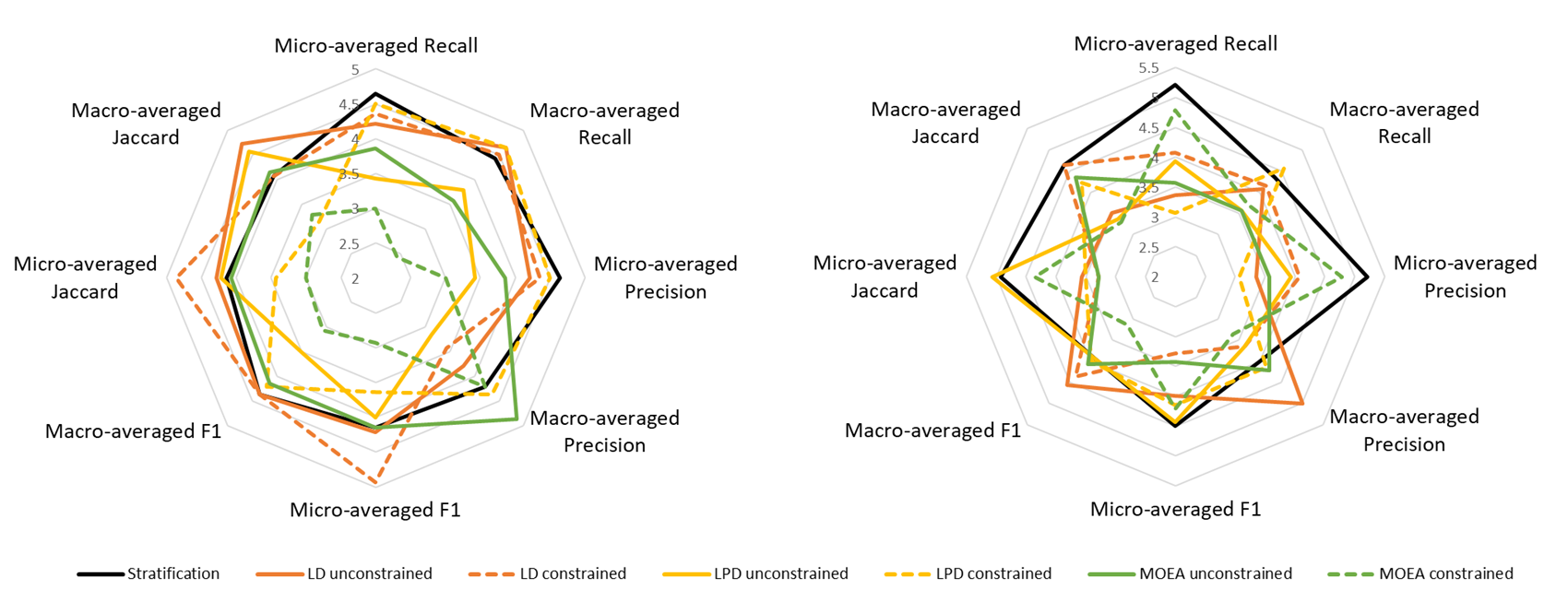}
\caption{Average ranks of the different methods when classification is performed using Label Powerset. Left figure shows the ranks regarding classification metrics. Right figure shows the ranks regarding the variance over folds.}
\label{img:LabelPowerset}
\end{figure}

Following
~\cite{sechidis2011stratification,szymanski2017network}, the results are discussed based on the average ranking of the different methods. The best method for a particular measure obtains a rank of 1, the next one a rank of 2, and so on. Figures~\ref{img:BinaryRelevance}~and~\ref{img:LabelPowerset} show the evaluations for both classification algorithms, BR and LP. Evolutionary approaches obtain, in general, better classification results than the Stratification method for all the metrics (left part of the figures). Even in the case of the constrained approaches, in which it is supposed to favor better results in macro-averaged metrics maybe in prejudice of micro-averaged metrics, evolutionary approaches behave better than Stratification. Using BR the best classification metrics are obtained by the unconstrained EA using LPD as fitness function followed by the constrained MOEA. Regarding variance over folds, these two methods are also the best ones. Using LP the best method is, for almost all the metrics, the constrained MOEA followed by the unconstrained EA using LPD. In terms of variance over folds, all the evolutionary approach have globally a similar behavior, being better than Stratification.

\section{Application to large image data sets}
\label{sec:large_datasets}

Given the recent relevance of multi-label computer vision applications using deep learning techniques, EvoSplit has also been validated using some large multi-label data sets widely employed by the research community: Microsoft COCO, ImageNet, and OpenImages. 

\subsection{Microsoft COCO}

The Microsoft Common Objects in COntext (MS COCO) data set~\cite{COCO} contains
$91$ common object categories with $82$ of them having more than $5,000$ labeled
instances. In total, the data set has $2,500,000$ labeled instances in $328,000$
images. This work has used the subset considered in the COCO Panoptic Segmentation Task, which includes $123,167$ images. From these, $5,000$ are selected for validation, and the remaining ones for training. The panoptic segmentation task involves assigning a semantic label and instance id for each pixel of an image, which requires generating dense, coherent scene segmentations. From all the data sets considered in this work, MS COCO is the one with the highest cardinality, an average of more than 11 labels per image. It is the only data set in which examples are not labeled stating if a label appears or not in an image but with the number of times that the label appears. As shown in Table~\ref{tab:measures}, this makes possible that the label (class~1~=~person) appearing the most in the data set (Max Frequency) does it more than once, on average, in each image. 

\subsection{Tencent ML-Images}

The Tencent ML-Images database~\cite{wu2019tencent} is a multi-label image database with 18M images and 11K categories, collected from ImageNet~\cite{imagenet} and OpenImages~\cite{OpenImages}. After a process of removal and inclusion of images, and relabeling of the data set, $10,756,941$ images, covering $10,032$ categories, are included from Imagenet. From these, $50,000$ are randomly selected as validation set. Following a similar process, $6,902,811$ training images and $38,739$ validation images are selected from OpenImages, covering $1,134$ unique categories. Finally, these images and categories from ImageNet and OpenImages are merged to construct the Tencent ML-Images database, which includes $17,609,752$ training and $88,739$ validation images ($50,000$ from ImageNet and $38,739$ from OpenImages), covering $11,166$ categories. 

\subsection{Results}

The measures of the application of the different methods to split these data sets are shown in Tables~\ref{tab:COCO}~to~\ref{tab:OpenImages}. In almost all the cases (in bold), any evolutionary approach, either single-objective or multi-objective, either constrained or unconstrained, performs better than the official or stratified splitting methods. Similarly to the results shown for traditional smaller data sets, MOEA shows the best combined results for the Label Distribution and the Label Pair Distribution in almost all the cases. For the Microsoft COCO and the OpenImages data sets the results for those measures are improved by one or more orders of magnitude with respect to the official splits, i.e. those offered by the providers of the data set.  

% Table generated by Excel2LaTeX from sheet 'Hoja1'
\begin{table}[t]
  \centering
  \caption{Evaluation of the different splitting approaches of the Microsoft COCO data set. In bold the results that are better than the official method, usually employed in the literature. The results of the constrained alternative is not presented as the unconstrained approach obtains $FZ$ and $FLZ$ measures equal to zero.}
  \setlength{\tabcolsep}{3pt}
  \small
    \begin{tabular}{|l|r|r|r|r|r|r|}
\cline{2-7}    \multicolumn{1}{r|}{} & \multicolumn{1}{c|}{\multirow{2}[4]{*}{\textbf{Official}}} & \multicolumn{1}{c|}{\multirow{2}[4]{*}{\textbf{Random}}} & \multicolumn{1}{c|}{\multirow{2}[4]{*}{\textbf{Stratification}}} & \multicolumn{2}{c|}{\textbf{Evolutionary algorithm}} & \multicolumn{1}{c|}{\multirow{2}[4]{*}{\textbf{MOEA}}} \\
\cline{5-6}    \multicolumn{1}{r|}{} &       &       &       & \multicolumn{1}{c|}{\textbf{LD}} & \multicolumn{1}{c|}{\textbf{LPD}} &  \\
    \hline
    \textbf{LD} & \num{2,42E-04} & \num{5,06E-04} & \num{2,54E-04} & \bm{$3.10 \times 10^{-5}$} & \num{2,76E-04} & \bm{$1.51 \times 10^{-5}$} \\
    \hline
    \textbf{LPD} & \num{7,24E-06} & \num{1,28E-05} & \bm{$5.66 \times 10^{-6}$} & \num{1,22E-05} & \bm{$6.48 \times 10^{-6}$} & \bm{$3.96 \times 10^{-6}$} \\
    \hline
    \textbf{ED} & $0$     & $0$     & $934$   & $0$     & $0$     & $0$ \\
    \hline
    \textbf{FZ} & $0$     & $1$     & $0$     & $0$     & $0$     & $0$ \\
   \hline
    \textbf{FLZ} & $0$     & $1$     & $0$     & $0$     & $0$     & $0$ \\
    \hline
    \end{tabular}%
  \label{tab:COCO}%
\end{table}%

% Table generated by Excel2LaTeX from sheet 'Tables paper'
\begin{table}[t]
  \centering
  \setlength{\tabcolsep}{1pt}
  \footnotesize
  \caption{Evaluation of the different splitting approaches of the Imagenet subset in the Tencent ML-Images database. In bold the results that are better than the official method, usually employed in the literature. NOTE: The results for the Stratification method are not shown as no result was obtained after 10 days of processing. Therefore, $ED$ is not shown as it only deviates from zero for that method.}
    \begin{tabular}{|l|r|r|r|r|r|r|r|r|}
\cline{2-9}    \multicolumn{1}{r|}{} & \multicolumn{1}{c|}{\multirow{3}[6]{*}{\textbf{Official}}} & \multicolumn{1}{c|}{\multirow{3}[6]{*}{\textbf{Random}}} & \multicolumn{4}{c|}{\textbf{Evolutionary algorithm}} & \multicolumn{2}{c|}{\multirow{2}[4]{*}{\textbf{MOEA}}} \\
\cline{4-7}    \multicolumn{1}{r|}{} &       &       & \multicolumn{2}{c|}{\textbf{Label Distribution}} & \multicolumn{2}{c|}{\textbf{Label Pair Distribution}} & \multicolumn{2}{c|}{} \\
\cline{4-9}    \multicolumn{1}{r|}{} &       &       & \multicolumn{1}{c|}{\textbf{Unconst.}} & \multicolumn{1}{c|}{\textbf{Const.}} & \multicolumn{1}{c|}{\textbf{Unconst.}} & \multicolumn{1}{c|}{\textbf{Const.}} & \multicolumn{1}{c|}{\textbf{Unconst.}} & \multicolumn{1}{c|}{\textbf{Const.}} \\
    \hline
    \textbf{LD} & \num{2,04E-06} & \num{2,86E-06} & \bm{$1.33 \times 10^{-6}$} & \bm{$1.35 \times 10^{-6}$} & \bm{$9.17 \times 10^{-7}$} & \bm{$8.02 \times 10^{-7}$} & \bm{$1.04 \times 10^{-6}$} & \bm{$1.48 \times 10^{-6}$} \\
    \hline
    \textbf{LPD} & \num{4,21E-07} & \num{6,45E-07} & \bm{$3.34 \times 10^{-7}$} & \bm{$3.36 \times 10^{-7}$} & \bm{$2.23 \times 10^{-7}$} & \bm{$1.94 \times 10^{-7}$} & \bm{$2.58 \times 10^{-7}$} & \bm{$3.49 \times 10^{-7}$} \\
    \hline
    \textbf{FZ} & $1$     & $1$ & $1$     & $1$     & $1$     & $1$     & $1$     & $1$ \\
    \hline
    \textbf{FLZ} & $361$   & $495$ & $466$   & \bm{$36$}    & $415$   & \bm{$36$}    & $429$   & \bm{$19$} \\
    \hline
    \end{tabular}%
  \label{tab:imagenet}%
\end{table}%

% Table generated by Excel2LaTeX from sheet 'Tables paper'
\begin{table}[t]
  \centering
  \setlength{\tabcolsep}{1pt}
  \footnotesize
  \caption{Evaluation of the different splitting approaches of the OpenImages subset in the Tencent ML-Images database. In bold the results that are better than the official method, usually employed in the literature. NOTE: The results for the Stratification method are not shown as no result was obtained after 10 days of processing. Therefore, $ED$ is not shown as it only deviates from zero for this method.}
    \begin{tabular}{|l|r|r|r|r|r|r|r|r|}
\cline{2-9}    \multicolumn{1}{r|}{} & \multicolumn{1}{c|}{\multirow{3}[6]{*}{\textbf{Official}}} & \multicolumn{1}{c|}{\multirow{3}[6]{*}{\textbf{Random}}} & \multicolumn{4}{c|}{\textbf{Evolutionary algorithm}} & \multicolumn{2}{c|}{\multirow{2}[4]{*}{\textbf{MOEA}}} \\
\cline{4-7}    \multicolumn{1}{r|}{} &       &       & \multicolumn{2}{c|}{\textbf{Label Distribution}} & \multicolumn{2}{c|}{\textbf{Label Pair Distribution}} & \multicolumn{2}{c|}{} \\
\cline{4-9}    \multicolumn{1}{r|}{} &       &       & \multicolumn{1}{c|}{\textbf{Unconst.}} & \multicolumn{1}{c|}{\textbf{Const.}} & \multicolumn{1}{c|}{\textbf{Unconst.}} & \multicolumn{1}{c|}{\textbf{Const.}} & \multicolumn{1}{c|}{\textbf{Unconst.}} & \multicolumn{1}{c|}{\textbf{Const.}} \\
    \hline
    \textbf{LD} & \num{2,25E-04} & \num{9,03E-06} & \bm{$2.89 \times 10^{-6}$} & \bm{$2.45 \times 10^{-6}$} & \bm{$2.63 \times 10^{-6}$} & \bm{$2.79 \times 10^{-6}$} & \bm{$2.15 \times 10^{-6}$} & \bm{$2.59 \times 10^{-6}$} \\
    \hline
    \textbf{LPD} & \num{1,18E-06} & \num{1,34E-07} & \bm{$1.15 \times 10^{-7}$} & \bm{$1.18 \times 10^{-7}$} & \bm{$6.89 \times 10^{-8}$} & \bm{$7.22 \times 10^{-8}$} & \bm{$8.73 \times 10^{-8}$} & \bm{$9.06 \times 10^{-8}$} \\
    \hline
    \textbf{FZ} & $1$     & $1$ & $1$     & \bm{$0$}     & $1$     & \bm{$0$}     & $1$     & \bm{$0$} \\
    \hline
    \textbf{FLZ} & $9$     & $27$ & $25$    & \bm{$0$}     & $11$    & \bm{$0$}     & $21$    & \bm{$0$} \\
    \hline
    \end{tabular}%
  \label{tab:OpenImages}%
\end{table}%

With these data sets it is even clearer the effect of using the constrained approach, in which the goal is to include all the labels in every fold. The introduction of the constraint allows to obtain $FZ$ and $FLZ$ equal to zero for the OpenImages data set, while their values for the official split are $1$ and $9$ respectively. $FLZ$ is dramatically reduced for the Imagenet data set ($361$ vs $19$ using MOEA).

\section{Conclusion}
\label{sec:conclusion}

This paper presents EvoSplit, a novel evolutionary method to split a multi-label data set into disjoint subsets. Different proposals, single-objective and multi-objective, using diverse measures as fitness function, have been proposed. A constraint has also been introduced in order to ensure that, if possible, labels are distributed among all subsets. In almost all the cases, the multi-objective proposal obtains state-of-the-art results, improving or matching the quality of the splits officially provided or obtained with iterative stratification methods. The improvement of EvoSplit over previous methods is highlighted when applied to very large data sets, as those currently used in machine learning and computer vision applications.

Moreover, the introduction of the constrained optimization decreases the chance of producing subsets with zero positive examples for one or more labels. This should have an effect on the training as there will be fewer labels for which there are no training or validation examples. A very relevant result is that EvoSplit is able to find splits that fulfill the constrain without affecting too much the distribution of labels and label pairs. 

EvoSplit is able to obtain better distributions of the original data sets  considering the desired size of the subsets as a hard constraint, i.e.~ensuring that the Examples Distribution is equal to zero. This is not the case for the iterative stratification method. The subsets obtained with EvoSplit allow better classification results and reduce the variance in the performance across subsets/folds.

Only in the case of the Imagenet data set, the best results are not obtained by the multi-objective EA but by the single-objective EA optimizing the Label Pair Distribution measure. An explanation to this might be related to the relation in the diversity between labels and pair labels for this data set. This data set has a particular characteristic: there are almost eight times more different pair labels than different labels in the data set (see the Diversity measure in Tables~\ref{tab:measures}~and~\ref{tab:measures2}). For all the other data sets, the relation is close to one. Therefore, this larger proportion in the diversity of label pairs might have an influence, benefiting the optimization based only on label pairs.

In conclusion, EvoSplit supports researchers in the process of creating a data set by providing different evolutionary alternatives to split that data set by optimizing the distribution of examples into the different subsets. EvoSplit can, in the future, be extended to higher levels of relationship between labels, e.g. triplets, by implementing a many-objective evolutionary algorithm~\cite{li2015}.

\section{Availability of splits and code}
\label{sec:splits}

The splits obtained with EvoSplit for the different data sets employed in this paper are freely available at \url{https://github.com/FranciscoFlorezRevuelta/EvoSplit} for their use by the research community. The EvoSplit code will be also available at the same repository.
 
\bibliography{references.bib}
\bibliographystyle{abbrv}

\end{document}